\documentclass{bmvc2k}

\usepackage{booktabs}
\usepackage{array}
\hypersetup{
  pdftitle={Beyond Report Imitation: Clinically Aware Multi-Image Ultrasound Report Generation from Visible Evidence},
  pdfauthor={Yuchen Yang, Wang Xin, Lufan Wang, Yinghong Pan, Yujuan Feng, Yuqing Yang},
  pdfsubject={British Machine Vision Conference 2026},
  pdfkeywords={ultrasound report generation, multimodal learning, evidence grounding, clinical alignment}
}

\makeatletter
\renewcommand\BMVA@blfootnote[1]{%
\begin{NoHyper}
\renewcommand\thefootnote{}\footnote{#1\hfill\textsuperscript{*}Corresponding author.}%
\addtocounter{footnote}{-1}%
\end{NoHyper}
}
\makeatother

\title{Beyond Report Imitation:\\
Clinically Aware Multi-Image Ultrasound Report Generation from Visible Evidence}

\addauthor{Yuchen Yang}{3319733548@bupt.edu.cn}{1}
\addauthor{Wang Xin}{wangxin21@pumch.cn}{2}
\addauthor{Lufan Wang}{wlf@bupt.edu.cn}{1}
\addauthor{Yinghong Pan}{panyinghong@bupt.edu.cn}{1}
\addauthor{Yujuan Feng}{yujuanf15@bjut.edu.cn}{3}
\addauthor{Yuqing Yang}{yangyuqing@bupt.edu.cn}{1,*}

\addinstitution{%
State Key Laboratory of Networking and Switching Technology,
\\Beijing University of Posts and Telecommunications, Beijing, China
}

\addinstitution{%
Department of Ultrasound, State Key Laboratory of Complex Severe and Rare Diseases,
Peking Union \mbox{Medical} \mbox{College} \mbox{Hospital}, Chinese Academy of
\mbox{Medical} \mbox{Sciences} \& Peking Union \mbox{Medical} \mbox{College}, Beijing, China
}

\addinstitution{%
College of Computer Science, Beijing University of Technology,
Beijing, China
}

\runninghead{Yang et al.}{Beyond Report Imitation}

\newcommand{\ours}{\textsc{CAMEO}\xspace}

\newcommand{\usreport}{\textsc{USReport-Distilled}\xspace}
\newcommand{\uspref}{\textsc{USReport-Pref}\xspace}
\newcommand{\pubmedus}{\textsc{PubMedVision-US}\xspace}

\begin{document}

\maketitle

\begin{abstract}
Generating ultrasound reports from multiple images requires aggregating clinical evidence across views, yet archived key frames capture only part of the dynamic examination. Raw-report imitation is therefore misaligned with visual supervision: content that is clinically valid for the full examination may be unverifiable from the images available to a model. This gap creates a clinical behavior alignment problem. A model must preserve visible findings, avoid diagnostic reversals and unsupported completion, and not collapse into conservative templates. We propose \textbf{\textsc{CAMEO}}, a \textbf{C}linically \textbf{A}ware \textbf{M}ulti-image \textbf{E}vidence-grounded \textbf{O}rchestration framework for ultrasound report generation. Stage~I learns ultrasound visual-language primitives; Stage~II performs Cross-View Evidence Grounding by distilling trusted visible report points into multi-image QA and report-style supervision; and Stage~III performs Clinically Aware Preference Alignment using clinical-error-oriented preference pairs. From USReport, we construct \usreport with 17,670 evidence-grounded paired-image training instances and \uspref with 21,869 preference pairs; we additionally use 25,631 \pubmedus ultrasound instruction samples for domain adaptation and multi-image instruction tuning. On the primary \usreport benchmark, \ours improves over EchoVLM from 0.25 to 0.40 BLEU-1, 0.28 to 0.45 ROUGE-1, and 0.27 to 0.43 METEOR, while raising ClinicalScore from 55.02 to 74.20. These results underscore the value of evidence-grounded supervision, clinically aware alignment, and clinically structured evaluation for reliable ultrasound report generation. Code and project resources: \url{https://github.com/NiHaoWoJiaoYYC/CAMEO}.
\end{abstract}

\section{Introduction}
\label{sec:intro}

Ultrasound is safe, relatively inexpensive, real-time, and widely used in routine care. Its interpretation is inherently multi-view: clinicians integrate scanning planes, probe-dependent appearances, and complementary key frames before writing a report. Multi-image ultrasound report generation is therefore not a collection of independent frame captions, but a problem of aggregating evidence across views and expressing it in clinically bounded language. A useful model must determine which observations are visible, how they relate across views, and how strongly they support report-level statements.

Recent MLLMs make this task increasingly feasible. Medical instruction tuning and large-scale biomedical image-text data improve domain coverage~\cite{LLaVAMed_2023,HuatuoGPTVision_2024,EchoVLM_2025}, while multi-image MLLMs support comparison, coreference, and cross-image aggregation~\cite{MANTIS_2024,LLaVAOneVision_2024,LLaVANeXTInterleave_2024}. However, input capacity and domain coverage alone do not guarantee reliable ultrasound reports. A model may produce fluent report text from modality priors or common templates while failing to preserve the specific evidence visible in the archived images.

A central source of this reliability problem is partial visual observability. A clinical ultrasound report may incorporate dynamic scanning, unarchived views, measurements, and contextual decisions, whereas a model usually receives only archived key frames. Some report statements are visible in these images; others may be clinically valid for the full examination but unverifiable from the available image set. Direct raw-report imitation therefore turns unobserved examination content into apparent image-grounded supervision.

\begin{figure*}[t]
\begin{center}
\includegraphics[width=0.96\linewidth]{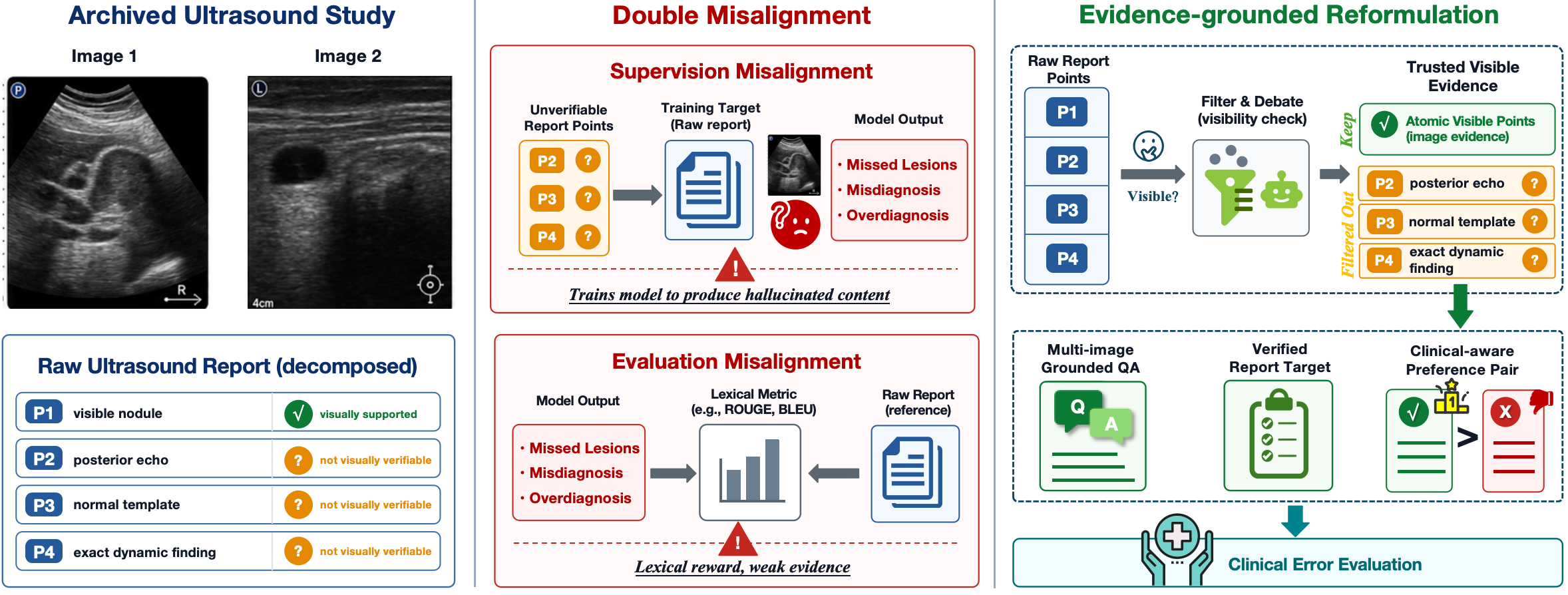}
\end{center}
\caption{Motivation. Archived ultrasound key frames only partially observe the dynamic examination. Raw report imitation can therefore introduce unsupported supervision, and reference-based metrics can reward non-grounded completion. \ours uses trusted visible evidence for cross-view training, preference alignment, and clinical error evaluation.}
\label{fig:motivation}
\end{figure*}

This mismatch creates the twofold misalignment illustrated in Fig.~\ref{fig:motivation}. During training, unverifiable report points can become positive visual targets, encouraging completion from language priors. During evaluation, lexical metrics may reward plausible but unsupported completion while penalizing evidence-faithful outputs that omit unverifiable reference text. This is not ordinary label noise: a reference report may be clinically correct for the full examination yet still be an inappropriate image-grounded target for the archived image pair.

The central challenge is not merely to remove unverifiable report content, but to define the desired model behavior under incomplete visual evidence. A reliable model should report visible abnormalities, preserve view-specific attribution, avoid unsupported diagnostic escalation, qualify claims when evidence is insufficient, and avoid conservative templates that omit visible findings. These requirements make report generation a clinically aware alignment problem rather than a data-cleaning exercise: the model must be informative when evidence is present and restrained when it is absent.

We formulate the task as \emph{clinically aware evidence-grounded generation under partial visual observability}. \ours distills trusted visible evidence from raw reports, converts it into cross-view SFT data and clinical-error-oriented preference pairs, and evaluates generations against the same evidence structure. Training proceeds through Stage I Ultrasound Domain Adaptation, Stage II Cross-View Evidence Grounding, and Stage III Clinically Aware Preference Alignment.

Our contributions are:
\begin{enumerate}
\setlength{\itemsep}{0pt}
\setlength{\parskip}{0pt}
\setlength{\parsep}{0pt}
\setlength{\topsep}{0pt}
\setlength{\partopsep}{0pt}
\item We formulate multi-image ultrasound report generation as clinically aware evidence-grounded generation under partial visual observability.
\item We introduce visual-evidence distillation, which decomposes raw reports into atomic points and retains visually supportable findings as positive image-grounded supervision.
\item We propose \ours, a stage-wise clinically aware alignment framework that combines cross-view SFT with clinical-error-oriented preference alignment to balance evidence attribution, informative reporting, and clinically bounded generation.
\item We introduce an evidence-grounded, LLM-assisted Clinical Error Evaluation protocol that measures whether generated reports are informative, diagnostically consistent, and bounded by trusted visible evidence.
\end{enumerate}

\section{Related Work}
\label{sec:related}

\subsection{Radiology Report Generation and Grounded Medical Reporting}
Radiology report generation commonly uses encoder--decoder models, memory modules, knowledge bases, or LLM adapters, as in R2Gen~\cite{R2Gen_2020} and R2GenGPT~\cite{R2GenGPT_2023}. These methods improve fluency and report-style generation, but typically treat the paired reference report as direct image-grounded supervision. This assumption is reasonable when most reference content is observable in the input images, but becomes fragile for archived ultrasound, where the report may summarize information from unobserved portions of the scan.

Grounded medical reporting further links report phrases to local visual evidence~\cite{Santiesteban_2024_BMVC}. This direction is complementary to ours, but typically begins with report content assumed to be visually relevant. We ask an earlier question: is each raw-report statement supportable from the archived ultrasound image pair at all? Only statements that pass this verification are used for positive supervision, preference construction, or evidence-grounded evaluation.

\subsection{Medical MLLMs and Ultrasound Intelligence}
Medical MLLMs such as LLaVA-Med~\cite{LLaVAMed_2023} and HuatuoGPT-Vision~\cite{HuatuoGPTVision_2024} improve biomedical visual instruction following, while ultrasound-specific models such as EchoVLM~\cite{EchoVLM_2025} address sonographic domain shift. These models provide useful priors for ultrasound language and image understanding, but domain adaptation alone does not specify how a model should respond when the report contains information that cannot be verified from the available views.

KMVE~\cite{KMVE_USReport_2024} is the closest prior work on ultrasound report generation. It introduces USReport and uses knowledge-guided alignment to reduce the image--report feature discrepancy. We use USReport as source data but pursue a different objective: rather than aligning images directly with raw reports, we reconstruct visually supportable evidence, build clinical-error-oriented preferences, and evaluate generations against trusted visible evidence.

\subsection{Multi-image and Cross-view Reasoning}
MANTIS~\cite{MANTIS_2024}, LLaVA-OneVision~\cite{LLaVAOneVision_2024}, and LLaVA-NeXT-Interleave~\cite{LLaVANeXTInterleave_2024} improve comparison and aggregation across images. MMIU~\cite{MMIU_2024} further demonstrates that multi-image understanding cannot be reduced to single-image VQA\@. Ultrasound reporting, however, requires more than generic comparison. A finding may be visible in one view, corroborated by both views, or unsupported by the available pair despite appearing in the raw report.

Cross-view reasoning therefore becomes a problem of clinical evidence attribution. The model must determine whether a statement is supported by the first image, the second image, both images, or neither, and express the result in appropriate report language. Stage II targets this requirement rather than generic multi-image conversation.

\subsection{Clinical Evaluation and Preference Alignment}
Clinical metrics such as CheXbert~\cite{CheXbert_2020}, RadGraph~\cite{RadGraph_2021}, RadCliQ~\cite{RadCliQ_2023}, and GREEN~\cite{GREEN_2024} move beyond lexical overlap, but generally compare generated reports with reference reports. In our setting, those references may contain unverifiable statements, so even clinical reference matching can reward unsupported completion. We instead evaluate against trusted visible evidence and separate omissions, incorrect statements, and unsupported additions.

Preference optimization provides a direct mechanism for behavior alignment. DPO~\cite{DPO_2023} avoids explicit reward modeling, and MMedPO~\cite{MMedPO_2025} demonstrates the value of clinically meaningful negative examples. Our preferences target ultrasound-specific omissions, polarity errors, view mismatches, diagnostic overreach, unsupported templates, and conservative shortcuts, using the same clinical error axes as the evaluation.

\section{Method}
\label{sec:method}

\begin{figure*}[t]
\begin{center}
\includegraphics[width=0.98\linewidth]{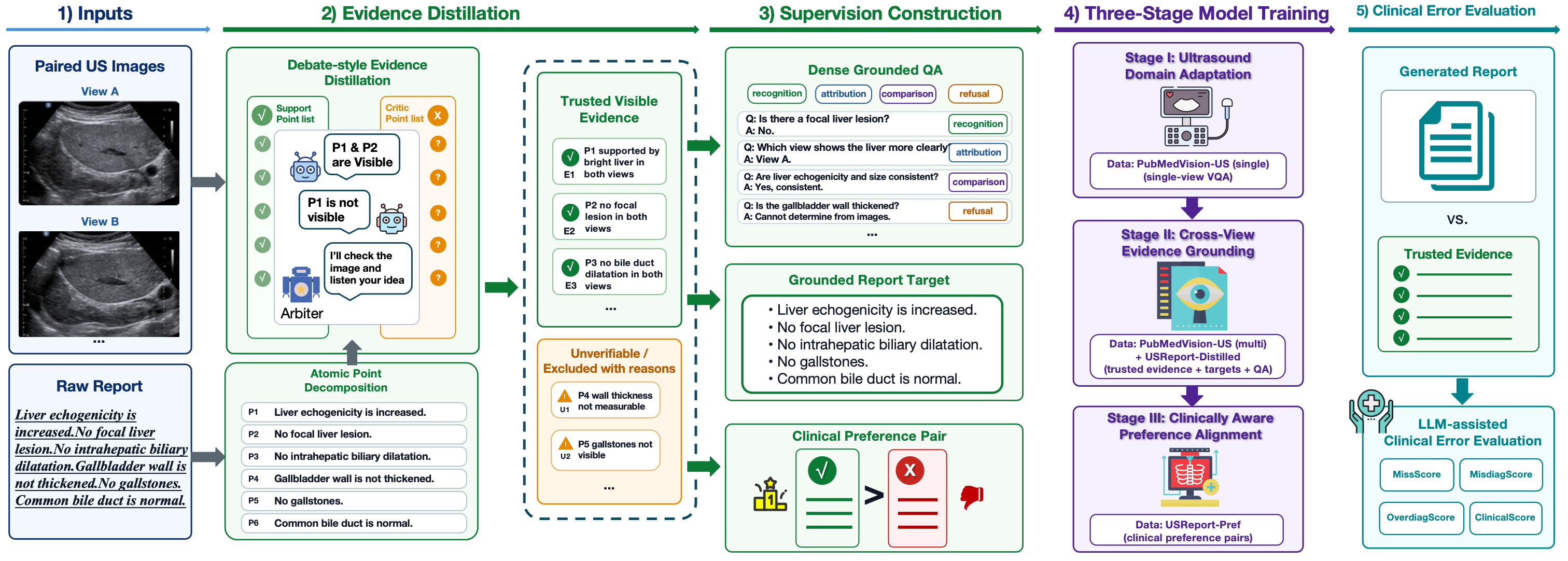}
\end{center}
\caption{Overview of \ours. Raw reports are decomposed into atomic clinical points and filtered into trusted visible evidence. The retained evidence supports grounded QA, report-style SFT, clinical preference pairs, and LLM-assisted Clinical Error Evaluation, aligning training and assessment within a shared evidence space.}
\label{fig:framework}
\end{figure*}

\subsection{From Partial Observability to Clinically Aware Evidence Grounding}
Let \(X=\{x_1,\ldots,x_N\}\) denote the archived ultrasound images and \(R\) the raw report. In our primary benchmark, \(N=2\). Unlike conventional report generation, which learns \(f_\theta:X\rightarrow R\), we do not assume that every statement in \(R\) is visually verifiable from \(X\). An unsupported statement may still be clinically valid for the full examination, but it should not serve as positive image-grounded supervision. This distinction frames partial observability as an evidence-allocation problem rather than an assertion that the original clinical report is incorrect.

We decompose \(R\) into atomic report points,
\begin{equation}
P=\{p_1,p_2,\ldots,p_M\},
\end{equation}
where each \(p_i\) expresses one factual clinical statement. For each point, we estimate a four-way visibility label
\begin{equation}
v_i\in\{\mathrm{visible},\mathrm{partial},\mathrm{unsupported},\mathrm{excluded}\},
\end{equation}
where \(\mathrm{partial}\) denotes a partially visible point. We define the scorable set and its fully and partially visible subsets as
\begin{equation}
\begin{aligned}
P_{\mathrm{scorable}} &= \{p_i\in P\mid v_i\neq\mathrm{excluded}\},\\
E_{\mathrm{vis}} &= \{p_i\in P_{\mathrm{scorable}}\mid v_i=\mathrm{visible}\},\\
E_{\mathrm{partial}} &= \{p_i\in P_{\mathrm{scorable}}\mid v_i=\mathrm{partial}\}.
\end{aligned}
\end{equation}
Each point also has a support source \(s_i\in\{\textrm{image 1},\textrm{image 2},\textrm{both},\textrm{none}\}\). Visible and partial points must identify at least one supporting view; \(s_i=\textrm{none}\) is reserved for unsupported or excluded points. The trusted positive image-grounded evidence is \(E_{\mathrm{vis}}\), and coverage is
\begin{equation}
\mathrm{coverage}(X,R)=\frac{|E_{\mathrm{vis}}|}{|P_{\mathrm{scorable}}|},
\end{equation}
with coverage set to zero when \(P_{\mathrm{scorable}}\) is empty. Partial points are included in the denominator but not the numerator; they support qualified supervision rather than fully positive report targets. Unsupported points remain scorable and motivate refusal or negative supervision, whereas excluded points enter neither the denominator nor downstream supervision. Coverage is a diagnostic signal rather than the task objective: low coverage indicates that report-level imitation is visually underdetermined, not that the case is uninformative. The reported training configuration does not discard cases solely because of low coverage. Coverage is retained as metadata for auditing this supervision; the complete policy is given in the supplementary material.

\subsection{Multi-source Ultrasound Data Organization}
Each data source has a distinct role. \pubmedus~\cite{HuatuoGPTVision_2024} provides broad ultrasound instruction data: its single-image subset supports Stage I, and its multi-image subset supports Stage II\@. USReport, released with KMVE~\cite{KMVE_USReport_2024}, provides paired-image reports for evidence distillation and preference construction. From USReport, we construct \usreport with 5,890 training cases, 11,780 images, 17,670 evidence-grounded instances, and 1,474 test cases. We also construct \uspref with 21,869 training and 1,354 validation preference pairs. Thus, \pubmedus supplies broad ultrasound instruction coverage, while the two derived datasets provide CAMEO's evidence and behavior supervision.

\subsection{Trustworthy Visual Evidence Distillation}
\paragraph{Atomic report point decomposition.}
We split each raw report into atomic points, each expressing one clinical fact, such as lesion presence, location, size tendency, echogenicity, boundary, wall appearance, polarity, or an assessable negative finding. Multi-organ, multi-lesion, and multi-attribute sentences are separated before verification so that a visible clause is not discarded merely because it appears alongside an unverifiable one. Negative findings are retained only when they are visually assessable from the archived views; otherwise, absence statements are treated as unsupported rather than as negative visual evidence.

\paragraph{Debate-style visibility verification.}
Two independent verifiers assign each point a visibility label, supporting view, visual cue, reason for non-support, and source span. The labels are \emph{visible}, \emph{partially visible}, \emph{unsupported}, and \emph{excluded}; support sources are image~1, image~2, both, or neither. An arbiter resolves disagreements over visibility, support source, and clinical interpretation. Only visible points become positive image-grounded supervision, whereas partially visible points support qualified responses. This procedure is scalable weak-supervision purification, not radiologist annotation.

\paragraph{Evidence reconstruction.}
Visible and partial points are reconstructed into a trusted evidence template containing key findings, optional visible findings, qualified partial findings, unsupported or excluded statements, support-image metadata, and coverage. The same template is used to generate SFT targets, construct preference contrasts, and evaluate model outputs. This shared representation keeps training and clinical evaluation anchored to a consistent definition of the available evidence.

\paragraph{Radiologist audit of evidence selection.}
LLM-generated supervision can fail silently: a fluent rationale need not reflect what is actually visible. We therefore assign distinct roles to the teacher models. Disagreements are resolved by the arbiter, after which schema validation and aggregation are deterministic. The resulting inclusion decisions are then subjected to blinded radiologist review. The reader saw only archived images and anonymized atomic points, without teacher labels or evidence templates. Across 301 annotations, the binary teacher labels achieved 78.9\% accuracy and 84.8\% F1 when the reader's ``supported'' and ``partially supported'' judgments were treated as positive. The reader also marked 25.2\% of scorable raw-report points as unsupported, directly quantifying the supervision noise introduced by raw-report imitation. The complete protocol and an independent output evaluation are provided in the supplementary material.

\subsection{Dense Evidence-grounded Supervision Construction}
We convert \(E_{\mathrm{vis}}\) and \(E_{\mathrm{partial}}\) into two complementary forms of supervision: fully visible points provide positive targets, while partial points provide explicitly qualified targets. SFT teaches multi-image reasoning, view attribution, and report-style expression, while preference data penalizes clinically plausible but unsafe behavior that likelihood training does not directly discourage. The model must learn both what the visible evidence supports and when fluent report completion would exceed that evidence.

\paragraph{SFT supervision from visible evidence.}
The SFT branch uses multi-image supervision from \usreport rather than imitating raw reports. The QA types in Table~\ref{tab:sft_supervision} target verification, attribution, cross-view synthesis, bounded refusal, and report-style generation. Verification separates learnable evidence from unverifiable text; attribution reduces view mismatch; synthesis teaches aggregation across complementary views; refusal teaches bounded generation when evidence is insufficient; and grounded-report QA converts trusted evidence into readable clinical language.

\begin{table*}[t]
\centering
\footnotesize
\setlength{\tabcolsep}{6pt}
\renewcommand{\arraystretch}{1.10}
\begin{tabular}{@{}>{\raggedright\arraybackslash}p{0.21\linewidth}>{\raggedright\arraybackslash}p{0.31\linewidth}>{\raggedright\arraybackslash}p{0.40\linewidth}@{}}
\toprule
USR-D QA type & Question / target form & Stage-II purpose \\
\midrule
Evidence QA & Is the report point visible in the paired images? & Filter out non-verifiable raw-report content. \\
View Attribution QA & Which view supports the finding: first view, second view, both, or neither? & Learn view-conditioned attribution and reduce cross-view confusion. \\
Cross-view QA & How do the two views jointly support or limit the finding? & Train reasoning over complementary views rather than independent single-image captioning. \\
Refusal QA & Should an unsupported claim be answered, qualified, or refused? & Teach conservative responses when the image pair is insufficient. \\
Grounded Report QA & Produce a concise report-style summary from trusted visible evidence. & Convert verified evidence into clinically readable report supervision. \\
\bottomrule
\end{tabular}
\caption{SFT supervision reconstructed from \usreport. Each QA type targets a distinct cross-view reporting behavior.}
\label{tab:sft_supervision}
\end{table*}

\paragraph{Preference data from clinical failure modes.}
The preference branch constructs \((y^+,y^-)\) pairs from the same evidence template. Rejected responses represent clinically plausible failure modes rather than generic low-quality text. After SFT, fluent normal templates or common diagnostic phrases may improve lexical overlap while omitting visible evidence, reversing attributes, or conflating views. Conversely, excessive refusal may reduce hallucination at the cost of under-reporting visible findings. The preferences therefore favor responses that are both informative and evidence-bounded, rather than merely shorter, more cautious, or more template-like.

Chosen responses cover visible findings, preserve attribution, and qualify incomplete support. Rejected responses instantiate organ-specific errors such as echogenicity reversal, boundary or wall-status confusion, benign-versus-suspicious wording, incorrect view attribution, unsupported normal templates, and conservative shortcuts. We group these errors under the same report-level axes used in evaluation: missed diagnosis, misdiagnosis, and overdiagnosis (Table~\ref{tab:pref_supervision}). The goal is evidence-bounded clinical informativeness, not shorter or more hesitant text.

\begin{table*}[t]
\centering
\footnotesize
\setlength{\tabcolsep}{4pt}
\renewcommand{\arraystretch}{1.10}
\begin{tabular}{@{}>{\raggedright\arraybackslash}p{0.16\linewidth}>{\raggedright\arraybackslash}p{0.17\linewidth}>{\raggedright\arraybackslash}p{0.34\linewidth}>{\raggedright\arraybackslash}p{0.25\linewidth}@{}}
\toprule
Preference contrast & Clinical axis & Rejected behavior & Desired alignment \\
\midrule
Omission & Missed diagnosis & A visible lesion, abnormality, or key descriptor is dropped. & Preserve clinically important visible evidence. \\
Unsupported content & Overdiagnosis & Template-like normal findings or organ status are added without visual support. & Suppress hallucinated and template-driven content. \\
Clinical attribute confusion & Misdiagnosis & Echogenicity, wall status, lesion boundary, or diagnostic tendency is reversed or confused. & Penalize fluent but clinically wrong statements. \\
View mismatch & Misdiagnosis & Findings are assigned to the wrong view or mixed across views. & Strengthen view-specific attribution and cross-view consistency. \\
Diagnostic overreach & Overdiagnosis & The conclusion is stronger than the paired images justify. & Encourage conservative language under partial observability. \\
Template shortcut & Missed diagnosis / Overdiagnosis & The response relies on fixed benign phrases or empty caution without reporting visible evidence. & Maintain clinical informativeness while staying evidence-grounded. \\
\bottomrule
\end{tabular}
\caption{Preference data from \uspref. Rejected responses instantiate clinically meaningful failure modes; chosen responses remain informative and evidence-bounded.}
\label{tab:pref_supervision}
\end{table*}

\subsection{Stage-wise Model Alignment}
Reliable multi-image ultrasound reporting requires domain adaptation, cross-view evidence grounding, and clinical risk calibration. We learn these capabilities sequentially because they address distinct sources of error: visual-language domain mismatch, multi-view evidence attribution, and generation behavior under partial evidence. Following LLaVA-Med~\cite{LLaVAMed_2023}, we keep the visual encoder fixed. Stage I begins with projector-only alignment and then instruction-tunes the projector and LLM; Stages II and III continue from the resulting checkpoint and update both the projector and LLM.

\paragraph{Stage I: Ultrasound Domain Adaptation.}
Stage I establishes ultrasound visual-language primitives. Projector-only alignment first maps ultrasound visual features into the language space while keeping the visual encoder and LLM fixed; single-image ultrasound instruction tuning then updates both the projector and LLM\@. This stage exposes the model to sonographic textures, acoustic artifacts, echogenicity descriptors, and organ-specific terminology. It reduces domain mismatch but does not address multi-image reporting, because view attribution and partial observability are not yet modeled.

\paragraph{Stage II: Cross-View Evidence Grounding.}
Stage II moves from ultrasound recognition to multi-image evidence use. It teaches which view supports a statement, whether the views jointly support it, and when insufficient evidence requires qualification. Multi-image ultrasound instructions provide cross-view practice, while \usreport supplies report-style targets derived from trusted visible evidence. At this stage, CAMEO learns to preserve visible findings and descriptors rather than defaulting to normal templates. We update the projector and LLM and optimize
\begin{equation}
\mathcal{L}_{\mathrm{SFT}}=-\sum_t \log \pi_\theta(y_t\mid X,q,y_{<t}).
\end{equation}

\paragraph{Stage III: Clinically Aware Preference Alignment.}
Stage III calibrates clinical behavior beyond likelihood training. \uspref contrasts preferred responses with alternatives containing unsupported completion, attribute or polarity errors, view mismatch, diagnostic overreach, and fixed-template shortcuts. This stage preserves the informativeness learned in Stage II while penalizing clinically plausible failures; it is behavior calibration rather than generic stylistic refinement. The aim is to reduce hallucinated or overly specific completion without driving the model toward empty refusals when visible evidence is present. We update the projector and LLM while keeping the reference model fixed. Given input \((X,q)\), preferred response \(y^+\), rejected response \(y^-\), and reference model \(\pi_{\mathrm{ref}}\), we optimize DPO~\cite{DPO_2023}
\begin{equation}
\mathcal{L}_{\mathrm{DPO}} =
- \log \sigma \left(
\beta \left[
\log \frac{\pi_\theta(y^+ \mid X,q)}{\pi_{\mathrm{ref}}(y^+ \mid X,q)}
-
\log \frac{\pi_\theta(y^- \mid X,q)}{\pi_{\mathrm{ref}}(y^- \mid X,q)}
\right]
\right).
\end{equation}
This increases the relative likelihood of evidence-bounded responses over clinically plausible failure cases.

\subsection{LLM-assisted Clinical Error Evaluation}
\label{sec:clinical_eval}
Conventional text metrics do not distinguish omissions from incorrect statements or unsupported additions. We therefore compare generated reports with the trusted evidence template. Here, \emph{missed diagnosis}, \emph{misdiagnosis}, and \emph{overdiagnosis} denote report-level textual errors rather than patient-level outcomes. The LLM judge emits structured semantic labels, from which all case-level scores are computed deterministically.

Let \(G\) denote trusted gold units, \(P_{\mathrm{clin}}\) non-style predicted units, and \(W_G=\sum_{g\in G}w_g\). Gold priorities use \(w_g\in\{3,2,1\}\) for key, visible, and optional findings. The judge assigns \(\ell_g\in\{0,0.5,1\}\) for matched, partially matched, and missed units. Wrong and unsupported-extra predictions receive subtype weights \(w_{\mathrm{wrong}}(p)\) and \(w_{\mathrm{extra}}(p)\) in \(\{1,2,3\}\); the complete mapping is provided in the supplementary material. The implemented losses are
\begin{align}
L_{\mathrm{miss}} &= \frac{\sum_{g\in G}w_g\ell_g}{W_G},\quad L_{\mathrm{miss}}=0\ \mathrm{if}\ W_G=0,\\
L_{\mathrm{mis}} &= \frac{\sum_{p\in P_{\mathrm{clin}}} w_{\mathrm{wrong}}(p)/3}{\max(|P_{\mathrm{clin}}|,1)},\quad
L_{\mathrm{over}} = \frac{\sum_{p\in P_{\mathrm{clin}}} w_{\mathrm{extra}}(p)/3}{\max(|P_{\mathrm{clin}}|,1)}.
\end{align}
Only units labeled wrong contribute to \(L_{\mathrm{mis}}\), and only units labeled extra contribute to \(L_{\mathrm{over}}\). If a unit is both wrong and unsupported, we count it primarily under \(L_{\mathrm{mis}}\). We report
\begin{align}
\mathrm{MissScore} &= 100(1-L_{\mathrm{miss}}),\quad
\mathrm{MisdiagScore} = 100(1-L_{\mathrm{mis}}),\\
\mathrm{OverdiagScore} &= 100(1-L_{\mathrm{over}}),\\
\mathrm{ClinicalScore} &=0.40\,\mathrm{MissScore}
+0.35\,\mathrm{MisdiagScore}
+0.25\,\mathrm{OverdiagScore}.
\end{align}
The aggregate weights prioritize omissions, followed by incorrect statements and unsupported completion. ClinicalScore is an expert-audited auxiliary proxy, not a clinical utility function; we report all three sub-scores to expose the underlying trade-offs.

\section{Experiments}
\label{sec:experiments}

\subsection{Experimental Setup}
\begin{figure*}[t]
\begin{center}
\includegraphics[width=0.98\linewidth]{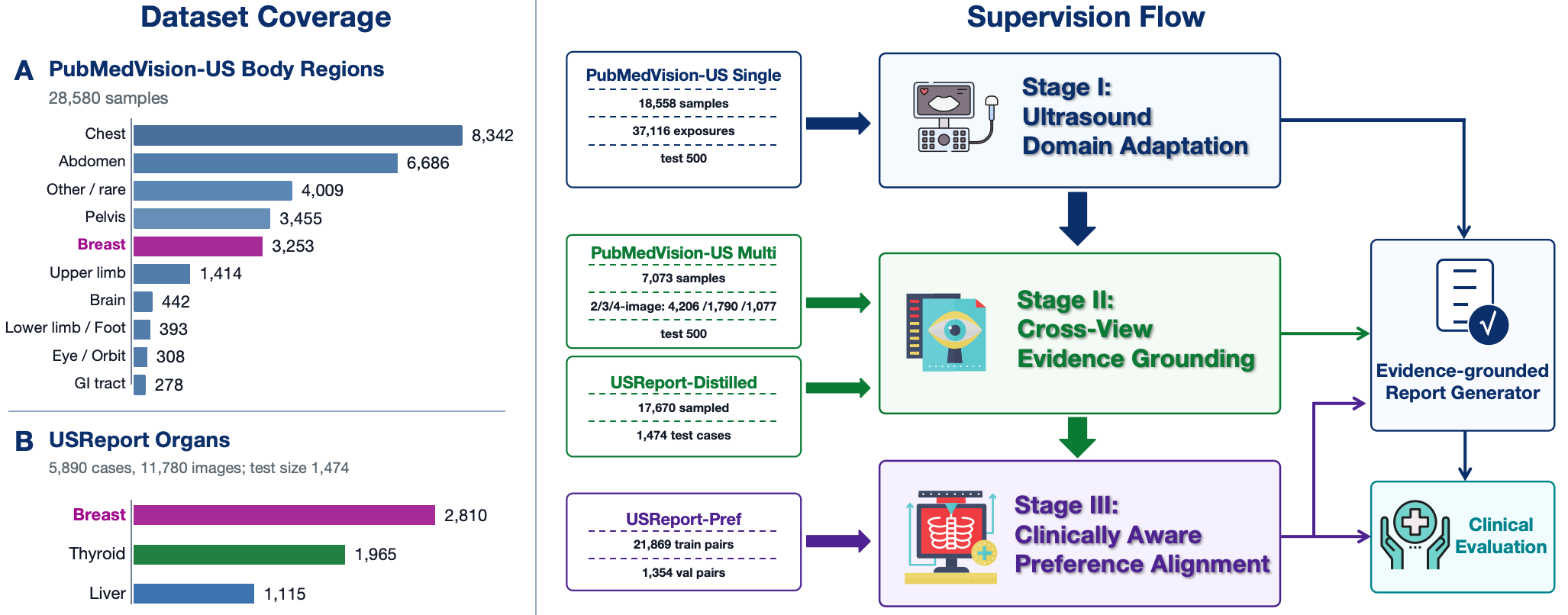}
\end{center}
\caption{Dataset coverage and supervision flow. \pubmedus~\cite{HuatuoGPTVision_2024} provides broad ultrasound instruction data, while USReport~\cite{KMVE_USReport_2024} is distilled into \usreport and \uspref for cross-view grounding, preference alignment, and clinical error evaluation.}
\label{fig:dataflow}
\end{figure*}

\paragraph{Datasets.}
Fig.~\ref{fig:dataflow} summarizes the role of each dataset. The \pubmedus{} single- and multi-image test sets each contain 500 samples. \usreport serves as the primary evidence-grounded benchmark, with 1,474 test cases. Its training split covers Breast, Thyroid, and Liver and contains 5,890 cases and 11,780 images. \uspref is used only for preference alignment and validation, never as a source of test references. This separation distinguishes broad ultrasound instruction coverage from the multi-image evidence and preference supervision used to train CAMEO.

\paragraph{Baselines and variants.}
We evaluate four external baselines. These are LLaVA-Med~\cite{LLaVAMed_2023}, Qwen3-VL-32B~\cite{Qwen3VL_2025}, Lingshu-32B~\cite{Lingshu_2025}, and EchoVLM~\cite{EchoVLM_2025}. Together, they cover medical instruction tuning, model scale, medical generalist pretraining, and ultrasound-specific modeling. The component analysis then isolates our evidence-grounded alignment using branches continued from Stage I, including Raw Report SFT and Distilled Report SFT\@. Dashes mark unavailable metrics, such as clinical scores for \pubmedus{} Multi.

All systems receive the same case-level multi-image inputs whenever their interfaces support them, together with a semantically consistent report-generation instruction. English outputs are scored directly; non-English outputs are translated only for text metrics, with clinical content preserved. Clinical metrics are computed by structured evidence-unit matching under the same rubric for every system. For \pubmedus{} Multi, EchoVLM was evaluated on the same 500 cases using matched English and Chinese prompts. Table~\ref{tab:main} reports the 500 English-prompt cases, and the supplementary material provides the paired Chinese-prompt control.

\paragraph{Implementation details.}
\ours is initialized from LLaVA-Med and trained in three stages on two modified NVIDIA RTX 4090 GPUs, each with 48GB of memory. The visual encoder remains frozen. Stage I performs projector-only alignment followed by joint projector--LLM instruction tuning; Stages II and III then update the projector and LLM using SFT and DPO, respectively. The teacher/debate pipeline operates offline and uses images plus raw reports solely to construct supervision. At inference, the 7B student receives only the images. Detailed model roles and settings are provided in the supplementary material.
\paragraph{Metrics.}
We report BLEU-1~\cite{BLEU_2002}, ROUGE-1/L~\cite{ROUGE_2004}, METEOR~\cite{METEOR_2005}, and BERTScore~\cite{BERTScore_2020}. On \usreport, the references are distilled evidence-grounded targets rather than raw reports; these metrics therefore quantify agreement with the target style, not independent clinical utility. We additionally report the MissScore, MisdiagScore, OverdiagScore, and ClinicalScore defined in Section~\ref{sec:clinical_eval}. Higher values are better for every metric.

\begin{table*}[t]
\centering
\scriptsize
\setlength{\tabcolsep}{3pt}
\renewcommand{\arraystretch}{1.10}
\resizebox{\linewidth}{!}{%
\begin{tabular}{llc ccccc cccc}
\toprule
Model & Type & Params & BLEU-1 & R-1 & R-L & METEOR & BERT & Miss & Misdiag & Overdiag & Clinical \\
\midrule
\multicolumn{12}{@{}l}{\textit{PubMedVision-US Multi}} \\
\addlinespace[1pt]
LLaVA-Med~\cite{LLaVAMed_2023} & medical & 7B & 0.19 & 0.30 & 0.20 & 0.23 & 0.19 & -- & -- & -- & -- \\
Qwen3-VL-32B~\cite{Qwen3VL_2025} & general & 32B & 0.32 & 0.38 & 0.23 & 0.34 & 0.17 & -- & -- & -- & -- \\
Lingshu-32B~\cite{Lingshu_2025} & medical & 32B & 0.33 & 0.42 & 0.25 & 0.35 & 0.25 & -- & -- & -- & -- \\
EchoVLM~\cite{EchoVLM_2025} & ultrasound & 12B & 0.12 & 0.19 & 0.13 & 0.14 & 0.08 & -- & -- & -- & -- \\
\ours & proposed & 7B & \textbf{0.38} & \textbf{0.44} & \textbf{0.33} & \textbf{0.43} & \textbf{0.30} & -- & -- & -- & -- \\
\addlinespace[3pt]
\midrule
\multicolumn{12}{@{}l}{\textit{USReport-Distilled}} \\
\addlinespace[1pt]
LLaVA-Med~\cite{LLaVAMed_2023} & medical & 7B & 0.17 & 0.21 & 0.16 & 0.16 & 0.18 & 22.35 & 65.28 & 69.64 & 49.19 \\
Qwen3-VL-32B~\cite{Qwen3VL_2025} & general & 32B & 0.26 & 0.33 & 0.24 & 0.36 & 0.28 & 25.25 & 70.39 & 73.26 & 53.05 \\
Lingshu-32B~\cite{Lingshu_2025} & medical & 32B & 0.25 & 0.30 & 0.21 & 0.30 & 0.25 & 26.31 & 69.35 & 75.48 & 53.66 \\
EchoVLM~\cite{EchoVLM_2025} & ultrasound & 12B & 0.25 & 0.28 & 0.21 & 0.27 & 0.20 & 29.28 & 77.41 & 64.87 & 55.02 \\
\ours & proposed & 7B & \textbf{0.40} & \textbf{0.45} & \textbf{0.34} & \textbf{0.43} & \textbf{0.42} & \textbf{56.36} & \textbf{82.90} & \textbf{90.53} & \textbf{74.20} \\
\bottomrule
\end{tabular}}
\caption{Main results on \pubmedus{} Multi~\cite{HuatuoGPTVision_2024} and \usreport, constructed from USReport~\cite{KMVE_USReport_2024}. R-1, R-L, and BERT denote ROUGE-1, ROUGE-L, and BERTScore. Clinical metrics are evaluated only on \usreport; the EchoVLM \pubmedus{} row uses all 500 English-prompt cases.}
\label{tab:main}
\end{table*}

\subsection{Main Results on Multi-Image Ultrasound Report Generation}
Table~\ref{tab:main} evaluates two settings. \pubmedus{} Multi measures transferable multi-image ultrasound generation using text metrics only, while \usreport is the primary evidence-grounded benchmark because it additionally supports clinical error metrics. On \pubmedus{} Multi, \ours reaches 0.38 BLEU-1, 0.44 ROUGE-1, and 0.43 METEOR, outperforming the external baselines under this protocol. This auxiliary result suggests that task-specific cross-view training benefits multi-image ultrasound instruction following. It is not, however, a general claim about ultrasound model quality, because the benchmark does not support clinical error evaluation.

On \usreport, Qwen3-VL-32B and Lingshu-32B achieve competitive lexical scores among the external baselines, yet their ClinicalScores are 53.05 and 53.66, respectively, showing that lexical plausibility and clinical evidence grounding can diverge. The ultrasound-specific EchoVLM attains a higher ClinicalScore than the larger general baselines but still trails \ours in this report-generation setting. Relative to EchoVLM, \ours improves BLEU-1 from 0.25 to 0.40, ROUGE-1 from 0.28 to 0.45, METEOR from 0.27 to 0.43, and ClinicalScore from 55.02 to 74.20. Ultrasound-specific modeling alone cannot replace task-specific cross-view evidence supervision and clinical alignment.

\paragraph{Independent radiologist evaluation.}
Automatic gains may reflect conformity to the evidence rubric rather than clinically meaningful report quality. We therefore asked one radiologist to blindly score four anonymized, randomly ordered outputs for each of 30 cases using only the images and reports. On a 1--5 scale, \ours scored $3.57{\pm}1.41$, compared with $1.73{\pm}0.98$ for EchoVLM, $1.90{\pm}0.99$ for Lingshu-32B, and $2.73{\pm}1.11$ for Raw Report SFT\@. The corresponding \ours win/tie/loss counts were 23/5/2, 23/7/0, and 22/3/5. ClinicalScore correlated with the radiologist's ratings of \ours (Spearman $\rho{=}0.698$, permutation $p{<}0.001$). In a separate 20-case preference audit, the radiologist selected the constructed chosen response in 17 cases (85\%; exact binomial $p{<}0.001$ against a $1/3$ chance rate). Together, these audits show that gains persist without access to the evidence schema, while residual losses expose remaining LLM failures. Full protocols for all three expert audits are provided in the supplementary material.

\subsection{Component Ablation Study}
Table~\ref{tab:ablation} isolates the role of each component. Stage I reduces ultrasound domain mismatch but remains insufficient without cross-view supervision. Raw Report SFT improves lexical overlap while retaining unverifiable content, reflecting the mismatch introduced by raw-report supervision. Distilled Report SFT improves target quality but lacks explicit paired-view attribution and consequently underperforms Stage I + Stage II\@. Stage II provides the largest gain by teaching cross-view evidence grounding. Stage III largely preserves text metrics while improving ClinicalScore, indicating that preference alignment primarily calibrates clinical behavior rather than lexical overlap.

\begin{table*}[t]
\centering
\footnotesize
\setlength{\tabcolsep}{3pt}
\renewcommand{\arraystretch}{1.10}
\resizebox{\linewidth}{!}{%
\begin{tabular}{lcccccc p{0.20\linewidth}}
\toprule
Variant & Evidence & Cross-view QA & Preference & BLEU-1 & ROUGE-1 & Clinical & Main purpose \\
\midrule
LLaVA-Med~\cite{LLaVAMed_2023} & no & no & no & 0.17 & 0.21 & 49.19 & base model \\
Stage I & no & no & no & 0.18 & 0.22 & 55.71 & domain adaptation \\
Stage I + Raw Report SFT & no & partial & no & 0.26 & 0.32 & 61.28 & raw imitation \\
Stage I + Distilled Report SFT & yes & limited & no & 0.36 & 0.41 & 63.01 & purification \\
Stage I + Stage II & yes & yes & no & 0.39 & 0.45 & 71.44 & cross-view reasoning \\
\ours & yes & yes & yes & 0.40 & 0.45 & 74.20 & clinical alignment \\
\bottomrule
\end{tabular}}
\caption{Component ablation study. Rows after Stage I continue from the Stage I checkpoint. Complete component metrics are provided in the supplementary material.}
\label{tab:ablation}
\end{table*}

\subsection{Clinical Error Analysis}
Table~\ref{tab:organ} breaks down the clinical scores by organ. Stage III raises the overall ClinicalScore by 2.76 points, with improvements along all three axes. Liver and thyroid improve consistently, suggesting that preference alignment reduces both incorrect statements and omissions when relevant structures are well represented in the archived views. Breast improves mainly in MisdiagScore; its essentially unchanged MissScore instead points to a remaining visual-recall bottleneck for subtle mammary findings. We therefore interpret the organ-level variation as an evidence-recognition limitation that preference alignment alone cannot resolve.

\begin{table}[!htbp]
\centering
\scriptsize
\resizebox{\linewidth}{!}{%
\begin{tabular}{lccc ccc ccc ccc}
\toprule
Organ & \multicolumn{3}{c}{Miss} & \multicolumn{3}{c}{Misdiagnosis} & \multicolumn{3}{c}{Overdiagnosis} & \multicolumn{3}{c}{Clinical} \\
\cmidrule(lr){2-4}\cmidrule(lr){5-7}\cmidrule(lr){8-10}\cmidrule(lr){11-13}
 & SFT & DPO & $\Delta$ & SFT & DPO & $\Delta$ & SFT & DPO & $\Delta$ & SFT & DPO & $\Delta$ \\
\midrule
Liver & 61.52 & 66.11 & +4.59 & 91.04 & 91.81 & +0.77 & 92.44 & 92.58 & +0.14 & 79.58 & 81.72 & +2.14 \\
Breast & 58.58 & 58.55 & -0.03 & 81.20 & 84.86 & +3.66 & 88.26 & 88.27 & +0.01 & 73.92 & 75.19 & +1.27 \\
Thyroid & 37.68 & 44.43 & +6.75 & 66.44 & 72.04 & +5.60 & 89.97 & 90.75 & +0.78 & 60.82 & 65.68 & +4.86 \\
All & 52.59 & 56.36 & +3.77 & 79.56 & 82.90 & +3.34 & 90.22 & 90.53 & +0.31 & 71.44 & 74.20 & +2.76 \\
\bottomrule
\end{tabular}}
\caption{Organ-wise clinical error evaluation. SFT denotes Stage II and DPO denotes \ours. Breast corresponds to the original Mammary label.}
\label{tab:organ}
\end{table}

\subsection{Qualitative Analysis and Limitations}
\begin{figure*}[!t]
\begin{center}
\includegraphics[width=0.98\linewidth]{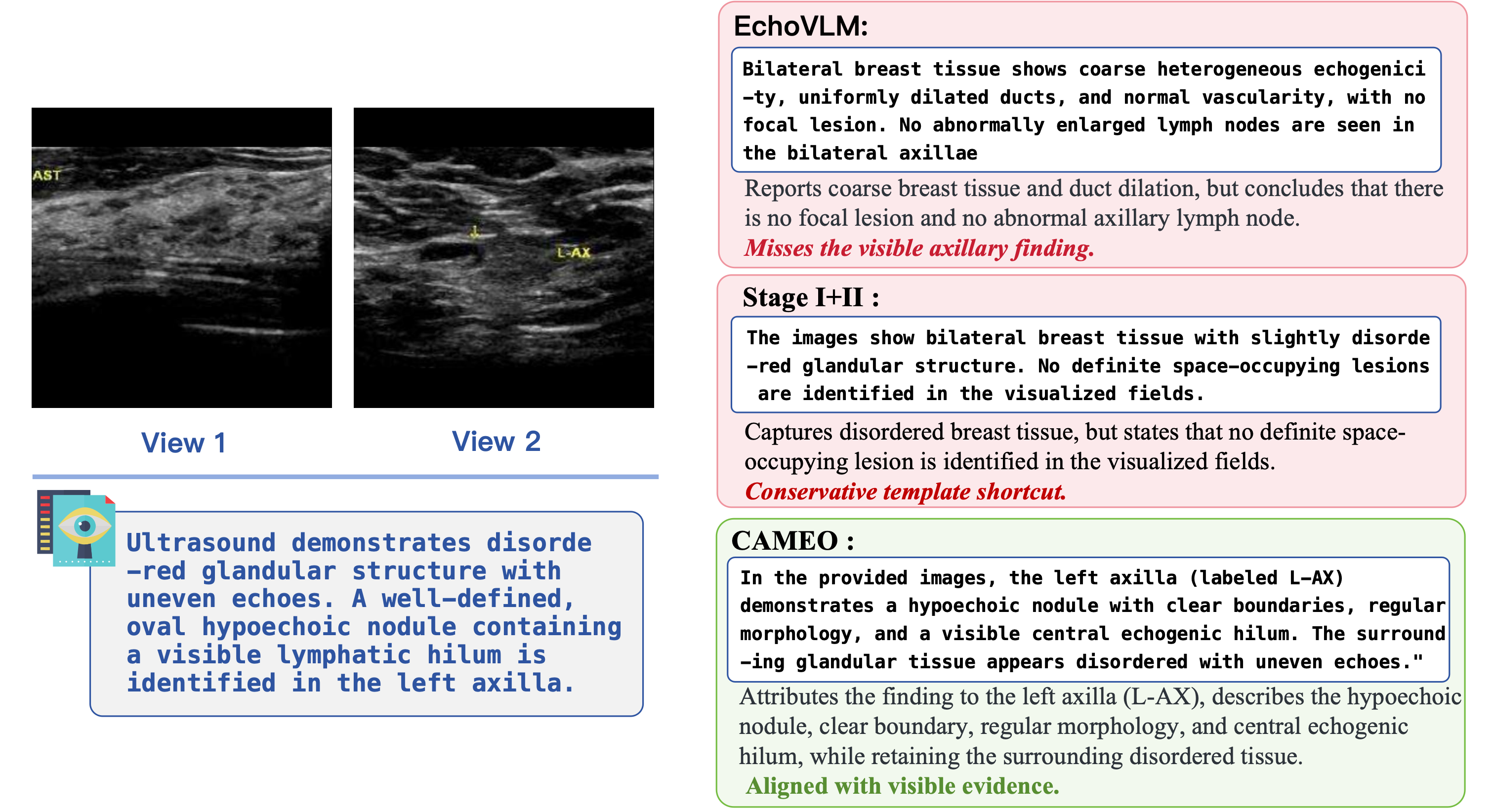}
\end{center}
\caption{Qualitative comparison on a breast/axillary case. EchoVLM and Stage I + Stage II omit the visible left-axillary nodule through normal or conservative templates, whereas \ours attributes the finding to the L-AX view and preserves the key descriptors.}
\label{fig:qualitative}
\end{figure*}

Fig.~\ref{fig:qualitative} illustrates the gap between report fluency and evidence grounding. EchoVLM produces a normal-style report and Stage I + Stage II remains conservative, both omitting the visible left-axillary nodule. \ours preserves the L-AX attribution and key descriptors without adding unsupported content. The example reflects the intended effect of Stage III: suppressing template shortcuts while preserving visible findings. It also remains consistent with Table~\ref{tab:organ}, where subtle mammary findings are still recall-limited.

This case also clarifies what CAMEO does not optimize for: the target is neither shorter output nor universal caution. A normal template may reduce unsupported abnormal claims while omitting visible evidence; an over-complete report may be informative but contain unsupported statements. The desired behavior is to preserve report-level informativeness when it can be anchored to view-specific evidence. Accordingly, our preference construction penalizes both unsupported completion and conservative shortcuts, and the clinical evaluation reports omissions and over-completion separately.

The quantitative results support the same interpretation. Stage I improves the base model's ultrasound language prior, but its gains are limited because single-image instruction tuning does not teach how evidence is distributed across views. Stage II delivers the largest report-generation improvement by converting trusted visible evidence into cross-view supervision. Stage III then reshapes the error profile rather than simply increasing text overlap: ClinicalScore improves while the lexical metrics remain similar, indicating that preference alignment primarily affects what the model includes, omits, or qualifies. This distinction matters in ultrasound reporting, where a small change in wording can alter clinical meaning through polarity, attribution, or diagnostic strength.

The baseline comparison further motivates evidence-grounded alignment. Strong general or medical VLMs can generate plausible report-like text, and ultrasound-specific pretraining supplies useful modality priors, but neither directly addresses the mismatch between raw reports and archived evidence. CAMEO makes this mismatch explicit throughout the pipeline: report points are filtered by visible support, SFT targets are tied to view-specific evidence, preference pairs encode clinically meaningful failures, and evaluation tests whether generated statements remain within the trusted evidence. Data construction, alignment, and evaluation are therefore treated as coupled design choices rather than independent components.

The radiologist analyses characterize, rather than eliminate, the limitations of current LLMs. They use a single reader and validate binary evidence inclusion, not four-class visibility, support-view attribution, or inter-reader reliability. Even \ours averages only $3.57{\pm}1.41$ on the 1--5 scale and loses to Raw Report SFT in 5 of 30 cases, revealing substantial case-level variation despite the aggregate improvement. The inputs are archived key frames rather than full dynamic examinations, and ClinicalScore remains an auxiliary proxy despite its correlation with reader ratings on the audited subset. The benchmark covers two-view report generation across three common organ categories, but not complete ultrasound protocols, dynamic scanning decisions, or rare findings. Our claims are therefore restricted to evidence-grounded report generation from archived images and do not extend to clinical deployment.

\section{Conclusion}
\label{sec:conclusion}
We introduced \ours, an evidence-grounded framework for multi-image ultrasound report generation under partial visual observability. It aligns visible evidence, supervision, generation, and evaluation through atomic-point distillation, cross-view tuning, and clinical-error-oriented preferences. Cross-View Evidence Grounding yields the largest gain, while preference alignment improves the clinical error profile.

The blinded radiologist studies support this design and expose its limits: raw reports contain image-unverifiable content, constructed preferences generally agree with expert judgment, and CAMEO improves report quality but retains case-specific failures. With one reader, archived key frames, three organ categories, and an auxiliary LLM-assisted metric, these results do not establish clinical utility.

\section*{Acknowledgements}
This research was supported by the National Natural Science Foundation of China (grants 62203060 and 62403492).

\bibliography{egbib}
\end{document}